# Unraveling Social Perceptions & Behaviors towards Migrants on Twitter

**Aparup Khatua, Wolfgang Nejdl**

L3S Research Center, Leibniz Universität Hannover, Hannover, Germany
khatua@l3s.de, nejdl@l3s.de

**Abstract**

We draw insights from the social psychology literature to identify two facets of Twitter deliberations about migrants, i.e., perceptions about migrants and behaviors towards migrants. Our theoretical anchoring helped us in identifying two prevailing perceptions (i.e., sympathy and antipathy) and two dominant behaviors (i.e., solidarity and animosity) of social media users towards migrants. We have employed unsupervised and supervised approaches to identify these perceptions and behaviors. In the domain of applied NLP, our study offers a nuanced understanding of migrant-related Twitter deliberations. Our proposed transformer-based model, i.e., BERT + CNN, has reported an F1-score of 0.76 and outperformed other models. Additionally, we argue that tweets conveying antipathy or animosity can be broadly considered hate speech towards migrants, but they are not the same. Thus, our approach has fine-tuned the binary hate speech detection task by highlighting the granular differences between perceptual and behavioral aspects of hate speeches.

## Introduction

The number of international migrants reached 272 million in 2020 compared to 150 million international migrants in 2010 (International Organization for Migration, & United Nations 2000). Our paper probes Twitter deliberations to explore the perceptions and behaviors towards these migrants. At the outset, we want to clarify that our paper cites some tweets that are offensive and vulgar towards migrants. However, we do not endorse the views expressed in these tweets but quote them only for academic research purposes. These offensive quotes do not reflect our opinions, and we strongly condemn offensive language on social media. Existing literature has labeled these offensive tweets as hate speeches (Davidson et al. 2017; Waseem and Hovy 2016). These hate tweets, and the propagation of hatred on the Twitter platform, elucidate the darker side of social media – which is a concern for society.

Generally, international immigrants come from economically weaker or politically disturbed countries, assuming that they would be leading a better life in the host countries. However, these migrants not only change the demographic fabric of the host nation but also impact the politics, law enforcement, economic, and labor market conditions in the host nation (Aswad and Menezes 2018). Consequently, a specific segment of the host countries can be apprehensive about these international immigrants. Recent political discourses during the 2016 and 2020 USA Presidential elections or Brexit referendum reveal an apprehensive view towards migrants (Khatua and Khatua 2016; Ogan et al. 2018; Waldinger 2018). On the contrary, the other segment of the society can be sympathetic towards migrants. This segment is concerned about the inequality and discrimination towards migrants. Social psychology literature argues that 'perception' mostly leads to 'behavior' (Dijksterhuis and Bargh 2001). According to this theory, apprehensiveness or antipathy towards migrants may lead to animosity or xenophobic behaviors. Similarly, a sympathetic view towards migrants may lead to solidarity. The scope of our paper did not allow us to investigate this causality. Hence, our paper attempts to understand the diverse and diametrically opposite migrant-related societal perceptions and behaviors on the Twitter platform. Based on the social psychology literature, we argue that understanding the discrimination faced by migrants or appreciating their struggles in asylums indicates a sympathetic *perception* of the user and getting involved in fundraising or support activities is a solidarity *behavior*. Similarly, assuming or believing that migrants are often involved with illegal activities is a negative *perception,* and demanding their deportation is a negative *behavior*. We have considered 0.8 million migration-related tweets (after pre-processing) from May 2020 to Sept 2020 to probe Twitter deliberations about migrants.

Opinion mining using social media data, especially in the context of migration, is a challenging task. To the best of our knowledge, none of the prior studies have analyzed the granular differences between perceptions and behavior towards migrants on social media platforms. Thus, in the domain of applied natural language processing (NLP), our study attempts to address this gap. We refer to interdisciplinary literature to conceptualize and identify perceptions and behaviors towards migrants on the Twitter platform. This is the core contribution of our study. Figure 1 reports our overall research framework.

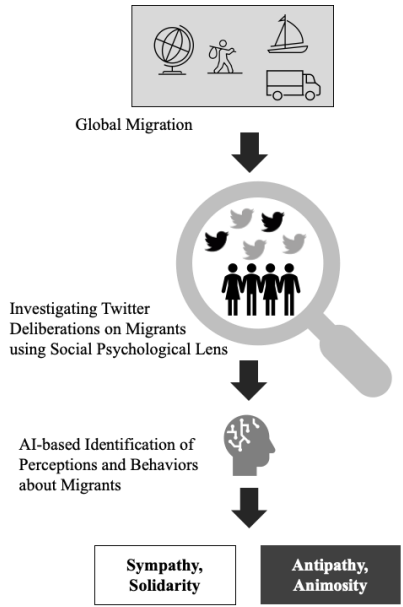

Figure 1: Flow of our Overall Research

A handful of prior studies probed the apprehensiveness towards migrants (which often get expressed through swear words or offensive languages). Literature has conceptualized this as a hate speech detection task (Davidson et al. 2017; Waseem and Hovy 2016). However, the literature mostly ignored the delicate nuances between perceptual and behavioral aspects of hate speeches. We argue that both types of tweets can be anti-migrant in their orientation, but they are not the same. Thus, we have reconceptualized the binary hate speech detection task into a fine-grained task of detecting the perceptual and behavioral aspects of hate speeches. This is another contribution of our study in the domain of applied NLP. Interestingly, we also note that a tweet that is supportive of migrants can also use 'swear words' against discrimination.

On the methodology front, we have employed unsupervised and supervised models for analyzing our corpus. We have employed three unsupervised, i.e., zero-shot classification models. Next, we consider convolutional neural network (CNN) and Bidirectional Long Short-Term Memory (Bi-LSTM) models with fastText embedding. Finally, we employ transformer-based models: Bidirectional Encoder Representations from Transformer (BERT) and Robustly Optimized BERT Pretraining Approach (RoBERTa) models. Our proposed BERT + CNN architecture has outperformed other models and reported an F1-weighted score of 0.76 for this complex perception-behavior identification task. To sum up, our paper tries to leverage AI for social good in the context of international migrants.

## Migration on Twitter: What we know?

Migration has attracted the attention of researchers from multiple disciplines, which range from sociology (Crawley and Skleparis 2018) to communication (Sajir and Aouragh 2019), and psychology (Goodman, Sirriyeh, and McMahon 2017; Volkan 2018) to information science (Aswad and Menezes 2018; Urchs et al. 2019; Vázquez and Pérez 2019). Migration-related issues were probed in the context of France (Siapera et al. 2018), Germany (Riyadi and Widhiasti 2020; Siapera et al. 2018), Italy (Capozzi et al. 2020; Kim et al. 2020b), Korea (Kim et al. 2020a), Netherlands (Udwan, Leurs, and Alencar 2020), Spain (Calderón, de la Vega, and Herrero 2020; Vázquez and Pérez 2019), Syria (Dekker et al. 2018; Rettberg and Gajjala 2016; Reel et al. 2018; Öztürk and Ayvaz 2018; Udwan, Leurs, and Alencar 2020), Turkey (Bozdag and Smets 2017; Özerim and Tolay 2020), the UK (Coletto et al. 2016), and the USA (Zagheni et al. 2018). Information science researchers mostly analyzed online contents, such as Facebook (Capozzi et al. 2020; Hrdina 2016; Zagheni et al. 2018), Instagram (Guidry et al. 2018), Pinterest (Guidry et al. 2018), YouTube (Lee and Nerghes 2018), Twitter (Alcántara-Plá and Ruiz-Sánchez 2018; Aswad and Menezes 2018; Calderón, de la Vega, and Herrero 2020; Gualda and Rebollo 2016; Kim et al. 2020; Nerghes and Lee 2018; Pope and Griffith 2016; Vázquez and Pérez 2019) as well as mainstream media (Nerghes and Lee 2019).

These studies have employed various NLP tools such as topic modeling (Calderón, de la Vega, and Herrero 2020; Guidry et al. 2018), sentiment analysis (Nerghes and Lee 2018; Öztürk and Ayvaz 2018; Pope and Griffith 2016), hashtag analysis (Özerim and Tolay 2020; Kreis 2017; Riyadi and Widhiasti 2020), and network analysis (Himelboim et al. 2017; Nerghes and Lee 2018; 2019).

The Twitter platform can be a real-time source of migration issues (Aswad and Menezes 2018). Hence, Twitter was widely employed by prior studies. Extant literature probed Twitter data to analyze migration movement (Mazzoli et al. 2020; Urchs et al. 2019; Zagheni et al. 2014). For example, Urchs et al. (2019) investigated the movement of migrants during 2015 in three European countries. They have identified 583 relevant tweets, which reveal the numbers of migrants moving from one country to another. Geo-tagged tweets were also used for analyzing migration movement (Mazzoli et al. 2020; Zagheni et al. 2014). Kim et al. (2020b) analyzed location information to identify immigrants and emigrants. Mazzoli et al. (2020) demonstrated that Twitter-based prediction of migration flow is consistent with official statistics. Coletto et al. (2016) considered spatial, temporal, and sentiment dimensions of their corpus and argued that Twitter provides real-time spatial information.

The Twitter data was also used for sentiment analysis and opinion mining in the context of migration (Lee and Nerghes 2018; Reel et al. 2018). A multilingual study (German and English) considered two specific refugee-related events and performed sentiment analysis of Twitter discussions around these two events (Pope and Griffith 2016). Similarly, Siapera et al. (2018) have analyzed various hashtags to study the network evolution as a response to three refugee-related specific events. This study argues that an event can have two predominant framings. First, a humanitarian frame where discussion revolves around how an organization can help refugees. Some of the prominent hashtags of this first frame were *#safepassage*, *#humanrights*, *#refugeesupport*. Second, a far-right perspective where refugees are framed as terrorists or criminals, and subsequently, these create security and safety concerns in the host country. These apprehensions towards migrants were also observed by other studies – especially in the context of Syrian refugees (Özerim and Tolay 2020; Öztürk and Ayvaz 2018; Reel et al. 2018). Reel et al. (2018) has proposed a random forest-based classifier to extract and identify tweets about Syrian refugees. Özerim and Tolay (2020) have explored Turkish tweets, especially against Syrian Refugees, and this study has observed the presence of echo chambers on the microblogging platform. Similarly, Kreis (2017) also analyzed negative perceptions about Syrian refugees through a hashtag-based analysis. These studies found a strong apathy towards refugees and noted nationalist hashtags such as *#EuropeforEuropeans*. However, Coletto et al. (2016) have found that positive and negative sentiments are not uniform in European Unions and emphasized the opinion dynamics. Khatua and Nejdl (2021) probed Twitter deliberations in the European context. They identified five themes, namely economic conditions, employment opportunities, healthcare support for migrants, discrimination against migrants, and safety concerns of the host countries.

Intuitively, nationalist ideologies may lead to abusive behaviors towards migrants. Davidson et al. (2019) found substantial racial biases in multiple hate speech and abusive language detection datasets. Detecting abusive online languages on social media platforms is a challenging task (Davidson et al. 2017; Waseem and Hovy 2016). We find that only a handful of studies, such as Calderón, de la Vega and Herrero (2020), i Orts (2019), Hrdina (2016), probed hate speeches in the context of migration. For example, the SemEval 2019 task tried to detect hate speech against immigrants (and women) on the Twitter platform (Basile et al. 2019). This task has two components: to detect the target of hate speech (generic or individual) and the presence (or absence) of aggressiveness. In the context of immigrants, Sanguinetti et al. (2018) has also prepared an Italian tweet corpus with binary labels for hate speech, stereotyping, and irony (i.e., yes, or no); multiple classes for aggressiveness and offensiveness (i.e., no, weak, and strong), and a five-point scale for intensity analysis. Similarly, Hrdina (2016) analyzed publicly visible pages and profiles on the Facebook platform. This study has found that hate speeches against migrants were aggravated by disparate Facebook users, extremist groups' propaganda, and news media. This study observed that frequent hate speech producers are primarily middle-aged and middle-class males and noted a significant under-representation of elderly and young Facebook users. Calderón, Blanco-Herrero, and Valdez Apolo (2020) also considered 1469 tweets to analyze the reasons behind the perception of rejecting migrants and refugees. This study argued that apathy towards foreigners is mainly driven by the economic burden of the host countries, security threat, invasion threat, identity threat, social prejudice, and explicit rejection. They also find that this rejection of foreigners is often due to multiple reasons from the above list.

To sum up, extant literature probed Twitter data for understanding latent opinions. Sentiment analysis, especially negative sentiment, reveals a xenophobic discourse on the social media platform. A handful of studies also employed an opinion mining approach to understanding the societal views about migrants and refugees. Most migration-related studies have adapted a syntactic approach for analyzing the linguistic content of their corpus. However, some recent studies have also considered advanced state-of-the-art neural models to decipher the semantic meaning. Yet, we did not come across an article that holistically investigated the diverse range of perceptions and behaviors towards migrants. Our research has attempted to address this gap, and we refer to social psychology literature to understand the perceptions and behaviors towards migrants.

## Perceptions and Behaviors towards Migrants

Social psychology literature argues that 'we perceive because we want to know what is going on around us ... perception is essential for us to comprehend our environment, but that does not mean that this understanding is an end in itself. Rather, understanding is a means by which we act effectively' (Dijksterhuis and Bargh 2001). This literature also assumes a 'shared representational systems for perception and action' because 'people have a natural tendency to imitate'. In other words, our perceptions and behaviors converge at the societal level due to our tendency to mimic others. Hence, societal perception is the cumulative outcome of individual perceptions. For example, the following anti-immigrant tweet from a specific user might be representing her personal view about immigrants.

- *We need to get rid of the Human Rights Act and political correctness. Immediately, we need to deport all illegal immigrants – the potential terrorists, plus those migrants convicted in criminal cases.*

However, like-minded social media users may propagate the above view by retweeting, and xenophobic behavior would gain momentum because we tend to imitate.

This perception-behavior theory has identified three trigger points of social perceptions. The first trigger point is *'observables'*, which is easy to understand because 'it involves behavior that we can literally perceive' (Dijksterhuis and Bargh 2001). Next, we develop *'trait inferences'* based on the behaviors of others. Interestingly, we generate trait inferences 'without being aware of it' (Dijksterhuis and Bargh 2001). Lastly, 'social perceivers also go beyond the information actually present in the current environment through the activation of *social stereotypes* (emphasis added) based on easily detectable identifying features of social groups' (Dijksterhuis and Bargh 2001). Cumulatively, we may perceive more than reality. For example, the following tweet captures a negative perception about migrants.

- *There is a high probability that a migrant has already committed a crime*

Intuitively, this perception was triggered either by *social stereotyping* or *trait interference*. The word 'probability' indicates that *observable* was not the trigger point for this perception. This *trait interference* or *social stereotyping*-based perception formation is crucial because without knowing the actual context, a specific segment of the society may develop an inappropriate perception about migrants. Our paper investigates Twitter deliberations to unravel these societal perceptions and behaviors towards migrants.

Theoretically, social psychology literature suggests that our perceptions about migrants might influence our behavior towards them (Ferguson and Bargh 2004). For instance, if we have sympathy towards migrants, then there is a high propensity that our behavior will express solidarity. On the contrary, antipathy may lead to animosity. However, it is worth noting that that the scope of our research did not allow us to investigate this causality between perception and behavior. At the user level, tracking an individual's tweets over a period and analyzing her perception and behavior for a sensitive issue like migration, even for academic purposes, can have ethical concerns. Testing the causality even at the societal level using Twitter data will be a challenging task. For instance, considering the evolution of Twitter deliberations over a longer time horizon might be an option; but this may broadly capture the composition of pro- versus anti-migrant users on the Twitter platform instead of the causality.

## Twitter Data and Annotation Process

**Data:** To explore perceptions and behaviors towards migrants, we have considered Twitter data and employed the Twitter search API (version Standard v1.1) for crawling data. Twitter API-based search allows to retrieve up to 1% of all the tweets on the Twitter platform. Morstatter et al. (2013) compared API-based data collection with Twitter's Firehose and found this API-based crawling 'is a sufficient representation of activity on Twitter as a whole'. Our initial crawling has considered keywords as follows: '*migrants*', '*refugee*', '*immigration*', and so on. We have regularly crawled data from May 2020 to September 2020. Prior migration-related studies observe that English tweets are predominant compared to other languages (Khatua and Nejdl 2021; Kim et al. 2020b). Accordingly, we also considered English tweets and crawled 1.2 million English tweets. We find a significant portion of our initial corpus was biased towards popular tweets. Hence, we have removed tweets/retweets with similar contents and duplicate tweet-ids. The corpus size became 0.8 million tweets after removing these popular tweets. Thus, 33% of our initial corpus (i.e., 0.4 million tweets out of a total of 1.2 million tweets) was repetitive tweets with similar contents.

**Geographical focus of our data:** The locational/country data is available for a small portion of these 0.8 million tweets. Based on this sub-sample, our corpus comprises tweets from 130 countries, but the distribution was skewed. Table 1 reports the geographical focus of our corpus. Around 75% of our corpus is from the USA, the UK, and Canada. One probably reason can be - English is the commonly used language in these countries.

| # | Country | Tweet (%) | # | Country | Tweet (%) |
|---|---------|-----------|---|---------|-----------|
| 1 | USA     | 53.6%     | 5 | Australia | 2.0%    |
| 2 | UK      | 18.4%     | 6 | Nigeria | 1.9%     |
| 3 | Canada  | 4.1%      | 7 | Others  | 16.1%    |
| 4 | India   | 3.9%      |   | Total   | 100.0%   |

Table 1: Geographical focus of our corpus

**Identification of Themes and Aspects:** Labelling our huge tweet corpus is a challenging task. Hence, Hedderich et al. (2020) suggest a distant and weak supervision approach for a new dataset. Here, a domain expert uses her tacit knowledge to design a set of rules based on contextual keywords (using external knowledge sources) and heuristics (Ratner et al. 2020; Rijhwani et al. 2020). However, this semi-automatic supervision approach, which is essentially syntactical, can lower the performance of classifiers (Fang and Cohn 2016). To overcome this limitation, prior studies suggest combining distant supervision with noise handling techniques (Hedderich et al. 2020). Following this stream of research, we have initially employed distant supervision and subsequently handled the noisy data through manual annotation. For designing our distant supervision rules, we have juxtaposed two threads of literature: perception-behavior studies and understanding of migration issues.

Perception is lexically defined as an idea or a belief you have based on how you see or understand something. Similarly, the lexical definition of behavior is 'the way that somebody behave, especially towards other people'. Subsequently, for distant supervision, we need to prepare an exclusive corpus of keywords - related to our categories of perceptions and behaviors. To prepare this corpus, authors have referred to multiple reports and scholarly articles on migration. The author also went through various United Nations High Commissioner for Refugees (UNCHR) policy documents to understand the context and identified the aspects accordingly.

Based on our understanding from the interdisciplinary literature, our *Sympathy Perception* comprises tweets expressing concerns about inequality, discrimination, and injustice towards migrants. Some of these tweets deliberate about discrimination in terms of low wages or inadequate facilities in the asylums. Thus, for identifying sympathy in our distant supervision approach, we have considered the following aspects: *vulnerable economic conditions*, *discrimination against migration*, *human rights violations*, *poor living conditions in asylums*, *lack of job opportunities*, *inadequate access to health/education facilities*. On the contrary, *Antipathy Perception* considers tweets that assume migrants are getting preferential treatment compared to citizens of host countries or most of them are involved in criminal or violent activities. Accordingly, for antipathy, we consider aspects such as *migrants enter illegally*, an *economic burden in host countries* (because they might destroy job opportunities for citizens of the host economy), *safety concerns* (because migrants can be violent out of desperation).

Our *Solidarity Behavior* tries to capture various support activities to rehabilitate the migrants. It ranges from fundraising activities to awareness campaigns. Thus, if social media users organize a donation drive and share the same on the Twitter platform, we consider it a solidarity behavior towards migrants. Aspects such as *support migrants/immigrants*, *donate for migrants*, the *safety of migrant women*, or *help refugee entrepreneurs* were considered under the solidarity category. On the contrary, the *Animosity Behavior* captures the disliking and hatred towards migrants. Hence, for animosity, we have considered the following aspects: *not in our country*, *no refugees*, *go back*, and *Europe for Europeans*.

We also find some migrant-related tweets which do not belong to the above four categories. These tweets are as follows: tweets that refer to migration superficially (where migration is not the dominant theme or core issue, but it might have an opinion about migrant issues) and tweets by migrants or refugees where they share their personal experiences. We label them as *Generic* tweets. The intersection between each category corpus and the tweet was computed to label a tweet in this weakly-supervised approach. We have considered this labeling based on the distant supervision approach as our silver standard (Ménard and Mougeot 2019). Subsequently, human annotators use their contextual understanding and domain knowledge to prepare the final gold standard by tackling the noisy data from the silver standard (Ménard and Mougeot 2019). The Cohen's Kappa coefficient was 82.8% for our inter-rater reliability.

**Complexity of our Task:** A fine-grained analysis of our corpus has elucidated the challenges associated with identifying perceptions and behaviors on the Twitter platform. We did not find much overlap between pro-migrant and anti-migrant categories, but we do observe overlaps within them – especially for anti-migrant tweets (i.e., antipathy and animosity categories). It is worth noting that social media platforms allow a user to express her views/voices. Hence, the puzzling question is - whether a voice is a perception or behavior? For example, let us consider two tweets as follows:

– *I find migrants to be a frightful lot, so different from us*
– *Hello, migrants! Go back to your own country*

Both the above tweets are anti-migrant tweets, but the first tweet is less provocative. Our annotation process has considered that the first tweet is 'a belief or opinion' and labeled it as an antipathy towards migrants. We felt that the second tweet is closer to 'the way that somebody behaves, especially towards other people' – hence, we considered it an animosity behavior.

However, the counter argument can be that migrants should go back to their own country – this can be 'a belief or opinion' and a significant portion of host country citizens might have the same opinion. Accordingly, the second tweet can be also considered as perception towards migrants. In other words, perception and behaviors are not dichotomous in a stringent sense, and we acknowledge this fluidity. In the context of anti-migrant tweets, perceptions are tweets which is less opinionated or less provocative.

We note that some tweets convey concerns from more than one of our four categories. Theoretically, a tweet can be a combination of any two or three (or even four) perceptions and behaviors. For instance, a tweet can be simultaneously pro- and anti-migrant as follows:

– *We must improve the living conditions of legal and needy migrants in our government asylums, but illegal migrants don't you f\*\*\*king dare to enter my country… you are criminal because you are entering illegally.*

To tackle these types of tweets, we need to frame our problem as a multi-label classification problem. However, our corpus doesn't have enough datapoints like the above tweet to train our neural network models. Hence, we ignored these tweets in our analysis, but future studies can probe these tweets. Additionally, some of those joint categories (e.g., sympathy + animosity) would be rare.

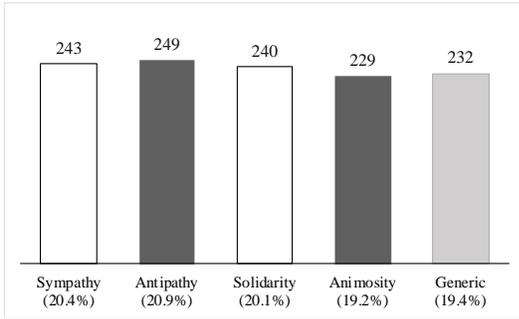

Figure 2: Distribution of Annotated Tweets

**Annotated Data:** Figure 2 reports the distribution of 1193 annotated tweets and Table 2 provides a few sample tweets from perception (i.e., sympathy & antipathy), behavior (i.e., solidarity & animosity) and generic categories. We paraphrased all quoted tweets in our paper to maintain user anonymity. We randomly split our 1193 annotated tweets into 85% (as training dataset) and 15% (as test dataset) for our subsequent analysis. Additionally, we also employed 5-folds cross-validation for our analysis.

| Class | Sample Tweets |
|---|---|
| Sympathy (SYM) | - Let us end stigma and discrimination against migrant workers and their children<br>- An asylum seeker is not an illegal immigrant … You f***ing idiot go buy a dictionary |
| Antipathy (ANT) | - It doesn't matter even though we were born here and pay for the healthcare. Just be a migrant and suddenly, it is a human rights violation<br>- We must stop the immigrants coming to our country they are crossing our borders in increasing numbers and putting the strain on our facilities |
| Solidarity (SOL) | - We believe that everyone deserves a fair chance to become an #entrepreneur. Therefore, we support #migrant entrepreneurs!<br>- Support our campaign today to help those in <location> facing all the ongoing humanitarian crises including the forgotten <location> refugees |
| Animosity (ANM) | - F**king illegal immigrants are not welcome in <location> F**k off you pr**k.<br>- You don't need no f**king answers. You are an immigrant and part of the problem. Just go back! |
| Generic (GEN) | - I am an immigrant and a citizen … I have paid taxes for 25 years and I care about this country<br>- Our data shows most <members of a political party> agree both that discrimination against whites has become as much of a problem as discrimination against immigrants |

Table 2: Representative tweets from our corpus

**Comparison with prior Hate Speech Corpora:** Anti-migrant tweets, which capture antipathy and animosity towards migrants, mostly use offensive languages or swear words. Thus, these tweets can be labeled as hate speeches, but they are not the same. Hence, these two classes deserve comparison with prior studies on hate speeches. A few prior studies considered voluminous annotated data, but some datasets were also smaller in size. For example, Ross et al. (2016) annotated 541 German tweets with key hashtags on the refugee crisis that could be offensive – this is comparable to our study. In comparison to prior studies, we find that our dataset is significantly complicated and balanced. For example, the dataset prepared by Davidson et al. (2017) contains 24,802 English tweets in English. However, only 5.77% of tweets were hate speech, 77.43% were offensive, and 16.80% were neither in these two categories. Similarly, Waseem and Hovy (2016) have considered 16914 English tweets. They have annotated this corpus into three classes: 12% tweets on racism, 20% tweets on sexism, and 68% tweets do not belong to either of these two classes. Madukwe, Gao, and Xue (2020) pointed out that these datasets were not balanced, which can inappropriately improve the classification accuracy. Unlike these prior studies, our dataset is balanced (refer to Figure 2).

Nowadays, Twitter allows its users to post 280 characters compared to the previous restriction of 140 characters. Hence, we find that the average length of annotated tweets of Davidson et al. (2017) is significantly shorter than our annotated tweets. For example, the average word count of their corpus is 14 without pre-processing, whereas the average word count of our corpus is 30 after pre-processing. Some of the annotated tweets from the corpus of Davidson et al. (2017) are as follows:

– *Bad b**ches is the only thing that I like.*
– *Foreign chick, no lie … Man, that b**ch beautiful.*

Intuitively, a less complex syntactic approach can correctly classify these shorter texts by considering context-specific swear words. However, longer tweets are more complex. For instance, a tweet from our corpus says:

– *We are bringing in thousands of migrants every year and they call us racist. No matter what we do or how much we give them these as*h***s will always view themselves as the oppressed.*

The above tweet indicates that the social media user has developed a negative perception and using offensive words towards migrants (probably) based on his experience. Another tweet from our corpus says:

– *If white Americans say, 'Take America back!' or tell an immigrant to 'Go back to your home country!', I am going to chuck a f**king history textbook in their face. White people originally came from Europe! Your f***ing ancestors were illegal immigrants!*

As we pointed out earlier, this tweet uses offensive words and argues that present American citizens' ancestors had moved from Europe to America. Hence, the legal citizens of the USA are historically immigrants. Therefore, the above tweet is sympathetic to today's immigrants, but a syntactic approach (by considering offensive words) will not be able to decipher it appropriately.

## Methodology and Findings

**Zero Shot Learning:** Building a rich training corpus is a time-consuming and resource-intensive task. Unsupervised models do not need this training corpus, but performances of unsupervised models are lower than supervised models. For instance, Nie et al. (2020) cautioned that non-expert annotators could successfully find the weakness of unsupervised models. However, it is worth noting that unsupervised models can perform the task without the training corpus. Hence, unsupervised methods, such as zero-shot learning models (ZSLM), are emerging as an alternate option in combination with large pre-trained models like BART (Lewis et al. 2019) and XLM-RoBERTa (Conneau et al. 2019). Hence, we consider unsupervised ZSLMs to predict unseen classes in the context of migration using the natural language inference (NLI) method. Yin, Hay, and Roth (2019) argue that pre-trained NLI models can perform the classification task without training. This approach trains a model to interpret the relationship (i.e., entailment, contradiction, or neutral) between two text streams. Next, it returns the probabilities of different classes according to their text content.

We have considered three pre-trained ZSLMs as follows: BART-Large-MNLI (Lewis et al. 2019), XLM-RoBERTa-Large-XNLI (Conneau et al. 2019), and XLM-RoBERTa-Large-XNLI-ANLI (Nie et al. 2020). BART-Large-MNLI considers a conventional seq2seq/machine translation architecture with a bidirectional encoder and a left-to-right decoder (Lewis et al. 2019). Using the MNLI dataset, this pre-trained model has shuffled the order of the original texts and employed an in-filling approach where a single mask token has replaced the spans of texts.

The training of XLM-RoBERTa-Large-XNLI has considered larger datasets, a more extensive vocabulary, and longer sequences with larger batches (Conneau et al. 2019). This model has considered 2.5 TB of newly created clean CommonCrawl data. This model is a combination of XLM and RoBERTa architecture. The approach makes full use of the entire content of the sentence to extract relevant semantic features. This model is the multilingual variant of RoBERTa, which has considered multilingual MLM for training. However, it also performs well for monolingual language task. Finally, XLM-RoBERTa-Large-XNLI-ANLI, took XLM-RoBERTa-Large as a base model, and fine-tuned it by combining NLI data by combining XNLI and ANLI across multiple languages (Nie et al. 2020). Recently, Nie et al. (2020) considered a new large-scale NLI benchmark dataset that was collected through an iterative, adversarial human-and-model-in-the-loop procedure.

**Results:** Table 3 reports the performance of transformer-based unsupervised models. We find that the weighted F1 score of unsupervised models for the single tag is significantly low, i.e., less than 0.30. Extant literature says that the NLI approach investigates the semantic similarity for predicting unseen classes. Hence, the tag word(s), specific to a class, can play a crucial role in the correct prediction. Pushp and Srivastava (2017) argued that multiple tag words could improve the accuracies of ZSLMs. Hence, we also followed a similar approach (refer to Table 3 for details). Our single tag approach considers only one keyword for each class (i.e., the word in Column 1 of Table 4). Double tags consider both the words of Column 2.

|  | PR | RC | F1 | AUC | T |
|---|---|---|---|---|---|
| BART-Large-MNLI (Lewis et al. 2019) | 0.35 | 0.34 | 0.27 | 0.58 | Single |
|  | 0.32 | 0.25 | 0.25 | 0.53 | Double |
|  | 0.43 | 0.41 | **0.38** | 0.63 | Multi |
| XLM-RoBERTa-Large-XNLI (Conneau et al. 2019) | 0.28 | 0.32 | 0.28 | 0.58 | Single |
|  | 0.25 | 0.28 | 0.25 | 0.55 | Double |
|  | 0.31 | 0.33 | **0.31** | 0.58 | Multi |
| XLM-RoBERTa-Large-XNLI-ANLI (Nie et al. 2020) | 0.31 | 0.33 | 0.27 | 0.58 | Single |
|  | 0.32 | 0.30 | 0.30 | 0.56 | Double |
|  | 0.36 | 0.35 | **0.31** | 0.59 | Multi |

*PR: Precision; RC: Recall; F1: F1-weighted; AUC: Area Under the ROC Curve; T: Tag Details; Best performance is shown in bold*

Table 3: Performance of Unsupervised Models

Interestingly, double tags have lowered the model performances, but multiple tags (i.e., Column 2 keywords + Column 3 keywords) have slightly improved model performances. Probably, double tags have created confusion (for closely resembling conceptual classes), whereas multiple tags have enhanced the interpretation of ZSLMs. Performances of these models are not impressive for our complex classification task, and these models failed to tease out the differences between broadly similar classes, such as antipathy and animosity. Additionally, ZSLM models wrongly classified some of the Generic class tweets into other classes. Our low Area Under the ROC Curve (AUC) scores also confirm the same. We find BART-Large-MNLI has reported the best F1-weighted score of 0.38. In Table 3, a few F1 scores are lower than precision or recall values because we have considered weighted F1-score.

| **Single tag** (Column 1) | **Double tags** (Column 2) | **Multi-tags** (Column 3) |
|---|---|---|
| Sympathy | Sympathy + Humanitarian | Empathy, Inequality |
| Antipathy | Antipathy + Xenophobic | Hatred, Disgust, Illegal |
| Solidarity | Solidarity + Consensus | Unity, Support |
| Animosity | Animosity + Bitterness | Deport, Hostility |
| Generic | Generic + Experiential | Impartial, Nondiscriminatory |

Table 4: Details of our tagging approach for ZSLMs

**Neural Models with Embedding:** Extant literature found that deep learning (DL) based classification models are superior to traditional bag-of-words models or n-gram models (Conneau et al. 2017; Kalchbrenner, Grefenstette, and Blunsom 2014; Young et al. 2018). For instance, the CNN model embeds words into low-dimensional vectors (Kim 2014). Next, convolutional filters slide over the word embedding matrix. These filters play a crucial role in task-specific performance. Finally, the max-pooling function provides a fixed dimension output for the desired classification task. In addition to CNN models, recurrent neural networks (RNN) were also used by prior studies for the classification task. However, RNNs cannot capture long-term dependencies of very long sequences. However, Bi-LSTM, which is a variation of RNN models, addresses this concern. Pre-trained embeddings improve the performance of these models. Hence, we have considered CNN and Bi-LSTM with pre-trained embeddings from fastText - wiki-news300d-1M, built using web-based corpus and statmt.org news dataset (Joulin et al. 2016).

|  | PR | RC | F1 | AUC | DR | BS |
|---|---|---|---|---|---|---|
| CNN + fastText | 0.71 | 0.69 | 0.70 | 0.81 | 0.4 | 16 |
|  | 0.75 | 0.68 | 0.70 | 0.82 | 0.5 | 16 |
|  | 0.75 | 0.71 | **0.72** | 0.83 | 0.4 | 32 |
|  | 0.71 | 0.69 | 0.70 | 0.81 | 0.5 | 32 |
| Bi-LSTM + fastText | 0.69 | 0.59 | 0.59 | 0.75 | 0.4 | 16 |
|  | 0.62 | 0.60 | 0.59 | 0.75 | 0.5 | 16 |
|  | 0.62 | 0.56 | 0.56 | 0.74 | 0.4 | 32 |
|  | 0.69 | 0.60 | **0.61** | 0.76 | 0.5 | 32 |

*PR: Precision; RC: Recall; F1: F1-weighted; AUC: Area Under the ROC Curve; DR: Dropout rate; BS: Batch size; Best performance is shown in bold*

Table 5: Performance of Deep Learning Models

**Results:** We have considered different hyperparameters, such as multiple batch sizes (16 and 32) and dropout rates (0.4 and 0.5), for our CNN and Bi-LSTM models. The hidden layer for all these models was 128. We have considered SoftMax activation in our final classification layer to predict the final class. We have considered 'adam' as our optimizer for the modeling. Table 5 reports the model performances. Performances of CNN + fastText models are better than Bi-LSTM + fastText models, and the highest F1-weighted Score for CNN is 0.72 (batch size 32, dropout 0.4).

**Transformer-based Neural Models:** Next, we have considered two transformer-based models: BERT (Devlin et al. 2019) and RoBERTa (Liu et al. 2019) for the classification task. Transformer-based models work reasonably well for text classification tasks because transformers are pre-trained on a diverse and large corpus. These models' core aspects are their multi-head self-attention to extract the input tokens' semantic aspects for contextual representation with multiple layers. Unlike RNNs, these models can handle long-term dependency problems. BERT has successfully performed numerous NLP-related tasks – including the classification task. BERT is a bidirectional unsupervised pre-trained model. Devlin et al. (2019) have considered BooksCorpus and English Wikipedia (16GB) for the training purpose. BERT was introduced in 2018. However, within a year, BERT's performance was further improved by adding more training corpus and incorporating minor adaptations to the training process (Liu et al. 2019). This advanced version of BERT is known as RoBERTa. In addition to the pre-training corpus of BERT, RoBERTa also used an additional corpus from CC-News (76 GB), Open Web Text (38 GB), and Storie's dataset (31 GB) for training.

Prior embedding approaches, such as word2vec (Mikolov et al. 2013) or GloVe (Pennington, Socher, and Manning 2014), have considered a single word embedding representation for each word without considering the context of that specific word. Therefore, these language representations failed to capture the context. In contrast, BERT considers the context of a particular word from both directions - both from the left and right direction. As we noted earlier, BERT and RoBERTa are pre-trained on a diverse and large corpus. This allows them to effectively understand most of the words used in online content compared to word2vec or GloVe. To sum up, BERT's fundamental principle is to employ bidirectional transformers for the feature extraction layer to extract the contextual meaning of the words. Following prior studies, we have fine-tuned our transformer-based models. BERT requires input data to be in a specific format. Thus, the [CLS] special token was used to indicate the beginning, and for the separation or the end of the sentence, the [SEP] was used. The next step was to tokenize the text corpus and extracting tokens that match BERT's vocabulary. For this task, we have used the HuggingFace python library (Wolf et al. 2020). This library includes pre-trained models and allows fine-tuning for the classification task. We have used '*BertForSequenceClassification*' for our classification task. We have considered the *BERT-Base-Uncased* model comprised of 12-layers and 12-heads with a total of 110M parameters. We have considered max_seq_length of 256.

As we mentioned earlier, the convolutional filters of CNN models play a crucial role in classification tasks. Thus, in combination with BERT, CNN can outperform $BERT_{Base}$ or BERT + LSTM (Dong et al. 2020; He et al. 2019; Mozafari, Farahbakhsh, and Crespi 2020). Accordingly, we consider the outputs from all individual layers of BERT architecture, and the outputs of each layer of the transformer are concatenated for the final result. We perform the convolutional operation with a window size - 3, hidden size of BERT – 768, and applying the max pooling on the convolution output from each transformer layer. Lastly, we concatenate these values, which is the input of the fully connected layer, before SoftMax performs the final classification task.

Additionally, we propose another BERT + CNN model that considers only the final layer of the BERT transformer for CNN-based classification. The dropout rate is 0.2 for all supervised models. For robustness, we have considered the following combinations of batch sizes (16, 32) and learning rates (1e -5, i.e., 0.00001, 2e-5, i.e., 0.0002, 3e-5, i.e., 0.00003). Like our DL models, we have considered 'adam' as our optimizer. Following prior studies, we have considered ten epochs for our *BERT-Base-Uncased* models. The open-source implementation, pre-trained weights, and full hyperparameter values and experimental details are in accordance with the HuggingFace transformer library (Wolf et al. 2020).

|  | PR | RC | F1 | AUC | LR | BS |
|---|---|---|---|---|---|---|
| BERT (Devlin et al. 2019) | 0.65 | 0.65 | 0.65 | 0.78 | 1e - 5 | 16 |
|  | 0.64 | 0.64 | 0.64 | 0.78 | 2e - 5 | 16 |
|  | 0.66 | 0.66 | 0.66 | 0.79 | 3e - 5 | 16 |
|  | 0.67 | 0.67 | **0.67** | 0.79 | 1e - 5 | 32 |
|  | 0.67 | 0.66 | **0.67** | 0.79 | 2e - 5 | 32 |
|  | 0.66 | 0.66 | 0.66 | 0.79 | 3e - 5 | 32 |
| RoBERTa (Liu et al. 2019) | 0.73 | 0.71 | 0.71 | 0.82 | 1e - 5 | 16 |
|  | 0.74 | 0.73 | **0.73** | 0.83 | 2e - 5 | 16 |
|  | 0.74 | 0.73 | **0.73** | 0.83 | 3e - 5 | 16 |
|  | 0.70 | 0.68 | 0.68 | 0.80 | 1e - 5 | 32 |
|  | 0.72 | 0.72 | 0.71 | 0.82 | 2e - 5 | 32 |
|  | 0.74 | 0.74 | **0.73** | 0.83 | 3e - 5 | 32 |
| Layer-wise BERT + CNN (Mozafari et al. 2020) | 0.73 | 0.70 | **0.71** | 0.81 | 1e - 5 | 16 |
|  | 0.69 | 0.69 | 0.69 | 0.81 | 2e - 5 | 16 |
|  | 0.71 | 0.71 | **0.71** | 0.82 | 3e - 5 | 16 |
|  | 0.65 | 0.65 | 0.64 | 0.78 | 1e - 5 | 32 |
|  | 0.69 | 0.67 | 0.68 | 0.80 | 2e - 5 | 32 |
|  | 0.67 | 0.67 | 0.67 | 0.80 | 3e - 5 | 32 |
| Final Layer of BERT + CNN (Proposed) | 0.74 | 0.74 | 0.74 | 0.84 | 1e - 5 | 16 |
|  | 0.79 | 0.76 | **0.76** | 0.85 | 2e - 5 | 16 |
|  | 0.77 | 0.74 | 0.75 | 0.84 | 3e - 5 | 16 |
|  | 0.72 | 0.70 | 0.71 | 0.82 | 1e - 5 | 32 |
|  | 0.73 | 0.73 | 0.72 | 0.83 | 2e - 5 | 32 |
|  | 0.75 | 0.73 | 0.72 | 0.83 | 3e - 5 | 32 |

*PR: Precision; RC: Recall; F1: F1-weighted; AUC: Area Under the ROC Curve; LR: Learning rate; BS: Batch size; Best performance is shown in bold*

Table 6: Performance of Transformer-based Models

**Results:** Table 6 reports the performances of transformer-based models. Our proposed BERT (final layer) + CNN architecture has outperformed other models. F1-weighted scores for some of the top-preforming models are as follows: 0.74 (95% confidence interval 0.67 – 0.81), 0.75 (95% confidence interval 0.68 – 0.82), and 0.76 (95% confidence interval 0.70 – 0.82. These are significantly high performances considering the complexity of our task due to closely resembling classes.

To test the efficiency of our proposed approach, we have also considered the publicly available dataset of Davidson et al. (2017). Interestingly, the F1-Score of our proposed BERT + CNN architecture is around 0.91 for the Davidson et al. (2017) dataset which is not balanced and comprises shorter texts with offensive words. This significant gap between 0.76 and 0.91 strongly indicates the complexity of our classification task.

**Error Analysis:** Table 7 reports a few wrongly classified tweets. We have highlighted (bold and italics) selected portions of these tweets to analyze why our model failed to classify these tweets correctly. For instance, syntactically, tweet #1 resembles an antipathy tweet, but semantically it is sympathetic. Our model has wrongly labeled tweet #5 due to the presence of the aspect such as 'infiltrating'. Similarly, the phrase 'I am an immigrant' in tweet #6 misguided our model. Our model labeled tweets #2 and #3 as generic due to presence of multiple issues, such as health, corruption, genocide, and so on, beyond immigration. In tweet #4, annotators felt the word 'criminal' conveys animosity. Similarly, tweet #7 used the word 'refugee' in a different context. In brief, the analysis of these wrongly classified tweets reveals the complexity of our task.

| # | Tweets from Test Dataset | ORI | PRED |
|---|---|---|---|
| 1 | Calling migrants as criminals when statistically ***illegal immigrants commit*** less ***violent crime*** than native born Americans is unfair! | SYM | ANT |
| 2 | I didn't get ***health benefits*** because I am a veteran and not an illegal immigrant. F*** you and f*** <location>. | ANM | GEN |
| 3 | F*** everyone in that parliament for turning a blind eye to illegal immigration, ***corruption, inequality, racism, genocide and trafficking***. | ANT | GEN |
| 4 | Hey, spineless <political party>, where is the social justice for the victims of illegal immigrant's crime. The same criminals you hide in your <location> | ANM | ANT |
| 5 | Let's do our part to prevent the gentrification <fast-food chains> from ***infiltrating*** <location> and other corridors. Support immigrant-owned businesses in <location> & <location> | SOL | ANT |
| 6 | True, ***I am an immigrant***, but this immigration numbers are outrages. Why do we need these many migrants? We don't have enough jobs in <location> | ANT | GEN |
| 7 | Try to understand the science of climate change - an environmental refugee ***is also a refugee*** in a broader sense! | GEN | SYM |

*OR: Original/True Label; PRED: Predicted Label; ANM: Animosity; ANT: Antipathy; GEN: Generic; SOL: Solidarity; SYM: Sympathy*

Table 7: Analysis of Wrongly Classified Tweets

## Conclusion

Prior studies from the information science domain have probed social media content to decipher the migration issues. Our literature review reveals that this literature has primarily investigated a focused issue or a specific topic. To the best of our knowledge, none of the prior studies holistically analyzed the social media deliberations. To address this gap, we draw insights from social psychology literature to identify various implicit perceptions and behaviors towards migrants. Figure 1 graphically presents our overall research framework. Our perception-behavior conceptualization of Twitter data is a contribution to the applied domain of NLP literature. This perception-behavior approach can potentially fine-tune future hate speech detection studies. Our proposed transformer-based supervised model, i.e., BERT + CNN architecture, has outperformed other models.

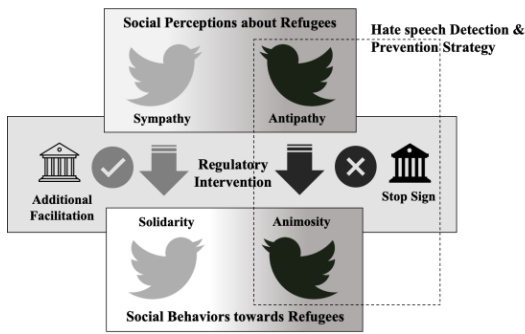

Figure 3: Takeaway for Policymakers

**Potential Policy Implications:** Interestingly, social psychology literature pointed out that perception mostly leads to action in animals, but there can be deviations for humans. Some perceptions may require an *'additional facilitating mechanism'* and 'sometimes the facilitator is present, sometimes it is absent; hence, sometimes perception leads to action whereas on other occasions it does not' (Dijksterhuis and Bargh 2001). The 'default option is that perception does lead to action (as in fish or frogs), but under some circumstances a "stop-sign" is given in order to block the impulse from resulting in overt behavior' (Dijksterhuis and Bargh 2001). Figure 3 presents the same graphically.

We argue that the above two possible roads of flexibility, either by using additional facilitating mechanisms or stop-sign, allow regulators to influence societal perceptions. For example, after identifying sympathetic perceptions, regulators can promote these perceptions by incorporating an '*additional facilitating*' mechanism such as endorsing or appreciating those sympathetic tweets. This will reinforce solidarity activities at the societal level. On the contrary, regulators can identify the negative (and mostly inaccurate) perceptions and use a 'stop sign' to weaken the antipathy-animosity link. For instance, a common misperception is that – inflow of migrants increases the labor supply. It lowers the wages and job opportunities – especially for the low-skilled employees of the host countries. However, in their recent book, Nobel laureates Banerjee & Duflo (2019), argued that migrants do not always lead to lower wages because these migrants 'spend money: they go to restaurants, they haircuts, they go shopping. This creates jobs, and mostly jobs for other low-skilled people'. Thus, regulators can also use social media to debunk the myth and counter inappropriate negative perceptions. This will reduce the animosity towards migrants and subsequently reduce xenophobic behaviors at the societal level. Twitter has a stringent policy against discrimination. Hence, regulators can also collaborate with Twitter to identify these inappropriate perceptions and label these tweets as disputed claims. Our research offers an AI-based framework to identify these societal perceptions. However, the implementation of these interventions is not within the scope of our study. Researchers from the communication domain need to perform psychological experiments to design appropriate and effective intervention mechanisms. Our study has tried to apply AI for social good. Hopefully, this study is an incremental step towards an egalitarian society.

## References

Alcántara-Plá, M., and Ruiz-Sánchez, A. 2018. Not for Twitter: Migration as a Silenced Topic in the 2015 Spanish General Election. In *Exploring Silence and Absence in Discourse*, 25-64.

Aswad, F.M.S., and Menezes, R. 2018, May. Refugee and Immigration: Twitter as a Proxy for Reality. In *FLAIRS Conference*, 253-258.

Banerjee, A.V., and Duflo, E. 2019. *Good Economics, Bad Economics: Six Ways We Get the World Wrong and How to Set It Right*. PublicAffairs.

Basile, V.; Bosco, C.; Fersini, E.; Debora, N.; Patti, V.; Pardo, F.M.R.; Rosso, P.; and Sanguinetti, M. 2019. Semeval-2019 task 5: Multilingual detection of hate speech against immigrants and women in Twitter. In 13th *International Workshop on Semantic Evaluation* (pp. 54-63). ACL.

Bozdag, C., and Smets, K. 2017. Understanding the images of Alan Kurdi with "small data": A qualitative, comparative analysis of tweets about refugees in Turkey and Flanders (Belgium). *International Journal of Communication*, 11,24.

Calderón, C.A.; Blanco-Herrero, D.; and Valdez Apolo, M.B. 2020. Rejection and Hate Speech in Twitter: Content Analysis of Tweets about Migrants and Refugees in Spanish. *Revista Española de Investigaciones Sociológicas*, 172, 21-40.

Calderón, C.A.; de la Vega, G.; and Herrero, D.B. 2020. Topic Modeling and Characterization of Hate Speech against Immigrants on Twitter around the Emergence of a Far-Right Party in Spain. *Social Sciences*, 9(11), 188.

Capozzi, A.; Morales, G.D.F.; Mejova, Y.; Monti, C.; Panisson, A.; and Paolotti, D. 2020. Facebook Ads: Politics of Migration in Italy. In *International Conference on Social Informatics*, 43-57.

Coletto, M.; Esuli, A.; Lucchese, C.; Muntean, C.I.; Nardini, F.M.; Perego, R; and Renso, C. 2016. Sentiment-enhanced multidimensional analysis of online social networks: Perception of the

Mediterranean refugees crisis. In *IEEE/ACM International Conference on ASONAM,* 1270-1277.

Conneau, A.; Khandelwal, K.; Goyal, N.; Chaudhary, V.; Wenzek, G.; Guzmán, F.; Grave, E.; Ott, M.; Zettlemoyer, L.; and Stoyanov, V. 2019. Unsupervised cross-lingual representation learning at scale. In *arXiv preprint arXiv:1911.02116.*

Conneau, A.; Schwenk, H.; Cun, Y.L.; and Barrault, L. 2017. Very deep convolutional networks for text classification. In *Proceedings of the 15th Conference of the European Chapter of the Association for Computational Linguistics*, 1, 1107-1116.

Crawley, H. and Skleparis, D. 2018. Refugees, migrants, neither, both: categorical fetishism and the politics of bounding in Europe's 'migration crisis'. *Journal of Ethnic and Migration Studies*, 44(1), 48-64.

Davidson, T.; Bhattacharya, D.; and Weber, I. 2019. Racial Bias in Hate Speech and Abusive Language Detection Datasets. In *Proceedings of the Third Workshop on Abusive Language Online*, 25-35

Davidson, T.; Warmsley, D.; Macy, M.; and Weber, I. 2017. Automated hate speech detection and the problem of offensive language. In *Proceedings of the International AAAI Conference on Web and Social Media*, 11,1.

Dekker, R.; Engbersen, G.; Klaver, J.; and Vonk, H. 2018. Smart refugees: How Syrian asylum migrants use social media information in migration decision-making. *Social Media+ Society*, 4(1).

Devlin, J.; Chang, M.W.; Lee, K.; and Toutanova, K. 2019. BERT: Pre-training of Deep Bidirectional Transformers for Language Understanding. *In NAACL-HLT (1).*

Dijksterhuis, A. and Bargh, J.A. 2001. The perception-behavior expressway: Automatic effects of social perception on social behavior. In *Advances in Experimental Social Psychology*. 33, 1-40.

Dong, J.; He, F.; Guo, Y.; and Zhang, H. 2020. A commodity review sentiment analysis based on BERT-CNN model. In *5th International Conference on Computer and Communication Systems (ICCCS)*, 143-147, IEEE.

Fang, M. and Cohn, T. 2016. Learning when to trust distant supervision: An application to low-resource POS tagging using cross-lingual projection. In *arXiv preprint arXiv:1607.01133*.

Ferguson, M.J. and Bargh, J.A. 2004. How social perception can automatically influence behavior. *Trends in cognitive sciences,* 8(1), 33-39.

Goodman, S.; Sirriyeh, A.; and McMahon, S. 2017. The evolving (re) categorisations of refugees throughout the "refugee/migrant crisis". *Journal of Community & Applied Social Psychology*, 27(2), 105-114.

Gualda, E. and Rebollo, C. 2016. The refugee crisis on Twitter: A diversity of discourses at a European crossroads. *Journal of Spatial and Organizational Dynamics*, 4(3),199-212.

Guidry, J.P.; Austin, L.L.; Carlyle, K.E.; Freberg, K.; Cacciatore, M.; Meganck, S.; Jin, Y.; and Messner, M., 2018. Welcome or not: Comparing# refugee posts on Instagram and Pinterest. *American Behavioral Scientist*, 62(4), 512-531.

He, C., Chen, S., Huang, S., Zhang, J., and Song, X. 2019. Using convolutional neural network with BERT for intent determination. In *International Conference on Asian Language Processing (IALP)*, 65-70, IEEE.

Hedderich, M.A.; Adelani, D.; Zhu, D.; Alabi, J.; Markus, U.; and Klakow, D. 2020. Transfer Learning and Distant Supervision for Multilingual Transformer Models: A Study on African Languages. In *Proceedings of the 2020 Conference on EMNLP*, 2580-2591.

Himelboim, I.; Smith, M.A.; Rainie, L; Shneiderman, B.; and Espina, C. 2017. Classifying Twitter topic-networks using social network analysis. *Social media+ society*, 3(1).

Hrdina, M. 2016. Identity, activism and hatred: Hate speech against migrants on Facebook in the Czech Republic in 2015. *Nase Spolecnost*, 1.

i Orts, Ò.G., 2019. Multilingual detection of hate speech against immigrants and women in Twitter at SemEval-2019 task 5: Frequency analysis interpolation for hate in speech detection. In *Proceedings of the 13th International Workshop on Semantic* Evaluation, 460-463.

International Organization for Migration, & United Nations. 2000. *World migration report*. Geneva, International Organization for Migration.

Joulin, A.; Grave, E.; Bojanowski, P.; Douze, M.; Jégou, H.; and Mikolov, T. 2016. Fasttext. zip: Compressing text classification models. arXiv preprint arXiv:1612.03651.

Kalchbrenner, N.; Grefenstette, E.; and Blunsom, P. 2014. A convolutional neural network for modelling sentences. In *52nd Annual Meeting of the Association for Computational Linguistics*.

Khatua, A. and Khatua, A. 2016. Leave or remain? Deciphering Brexit deliberations on Twitter. In *2016 IEEE 16th* ICDMW, 428-433, IEEE.

Khatua, A. and Nejdl, W. 2021. Analyzing European Migrant- related Twitter Deliberations. In *Companion Proceedings of the Web Conference 2021 (WWW '21 Companion)*.

Kim, Y. 2014. Convolutional neural networks for sentence classification. In *Proceedings of the 2014 Conference on EMNLP*, 1746–1751, ACL.

Kim, D.O.D.; Curran, N.M.; and Kim, H.T.C. 2020a. Digital Feminism and Affective Splintering: South Korean Twitter Discourse on 500 Yemeni Refugees. *International Journal of Communication*, 14,19.

Kim, J.; Sîrbu, A.; Giannotti, F.; and Gabrielli, L. 2020b. Digital footprints of international migration on Twitter. In *International Symposium on Intelligent Data Analysis,* 274-286, Springer.

Kreis, R. 2017. # refugeesnotwelcome: Anti-refugee discourse on Twitter. *Discourse & Communication*, 11(5), 498-514.

Lee, J.S. and Nerghes, A. 2018. Refugee or migrant crisis? Labels, perceived agency, and sentiment polarity in online discussions. *Social Media+ Society*, 4(3).

Lewis, M.; Liu, Y.; Goyal, N.; Ghazvininejad, M.; Mohamed, A.; Levy, O.; Stoyanov, V.; and Zettlemoyer, L. 2019. BART: Denoising sequence-to-sequence pre-training for natural language generation, translation, and comprehension. In *arXiv preprint arXiv:1910.13461*.

Liu, Y.; Ott, M.; Goyal, N.; Du, J.; Joshi, M.; Chen, D.; Levy, O.; Lewis, M.; Zettlemoyer, L.; and Stoyanov, V. 2019. RoBERTa: A Robustly Optimized BERT Pretraining Approach. In *arXiv preprint arXiv:1907.11692*.

Madukwe, K.; Gao, X.; and Xue, B. 2020. In Data We Trust: A Critical Analysis of Hate Speech Detection Datasets. In *Proceedings of the Fourth Workshop on Online Abuse and Harms*, 150-161.

Mazzoli, M.; Diechtiareff, B.; Tugores, A.; Wives, W.; Adler, N.; Colet, P.; and Ramasco, J. J. 2020. Migrant mobility flows characterized with digital data. *PloS one*, 15(3), e0230264.

Ménard, P.A. and Mougeot, A., 2019. Turning silver into gold: error-focused corpus reannotation with active learning. In *Proceedings of the International Conference on RANLP*, 758-767.

Mikolov, T.; Chen, K.; Corrado, G.; and Dean, J. 2013. Efficient estimation of word representations in vector space. In *arXiv preprint arXiv:1301.3781*.

Morstatter, F., Pfeffer, J., Liu, H., and Carley, K. M. 2013. Is the sample good enough? Comparing data from Twitter's streaming API with Twitter's firehose. In *7th International AAAI conference on weblogs and social media*.

Mozafari, M.; Farahbakhsh, R.; and Crespi, N. 2020. Hate speech detection and racial bias mitigation in social media based on BERT model. *PloS one*, 15(8).

Nerghes, A. and Lee, J.S. 2018. The refugee/migrant crisis dichotomy on Twitter: A network and sentiment perspective. In *Proceedings of the 10th ACM conference on web science*, 271-280.

Nerghes, A. and Lee, J.S. 2019. Narratives of the refugee crisis: A comparative study of mainstream-media and Twitter. *Media and Communication*, 7(2 Refugee Crises Disclosed), 275-288.

Nie, Y.; Williams, A.; Dinan, E.; Bansal, M.; Weston, J.; and Kiela, D. 2020. Adversarial NLI: A New Benchmark for Natural Language Understanding. In *Proceedings of the 58th Annual Meeting of the Association for Computational Linguistics*, 4885-4901.

Ogan, C.; Pennington, R.; Venger, O.; and Metz, D., 2018. Who drove the discourse? News coverage and policy framing of immigrants and refugees in the 2016 US presidential election. *Communications*, 43(3), pp.357-378.

Özerim, M.G. and Tolay, J. 2020. Discussing the populist features of anti-refugee discourses on social media: an anti-Syrian hashtag in Turkish Twitter. *Journal of Refugee Studies*.

Öztürk, N. and Ayvaz, S. 2018. Sentiment analysis on Twitter: A text mining approach to the Syrian refugee crisis. *Telematics and Informatics*, 35(1), 136-147.

Pennington, J.; Socher, R.; and Manning, C.D. 2014. Glove: Global vectors for word representation. In *Proceedings of the 2014 conference on empirical methods in natural language processing (EMNLP)*, 1532-1543).

Pope, D. and Griffith, J. 2016. An Analysis of Online Twitter Sentiment Surrounding the European Refugee Crisis. In *Proceedings of the International Joint Conference on Knowledge Discovery, Knowledge Engineering and Knowledge Management*, 299-306.

Pushp, P.K. and Srivastava, M.M. 2017. Train once, test anywhere: Zero-shot learning for text classification. In *arXiv preprint arXiv:1712.05972*.

Ratner, A.; Bach, S.H.; Ehrenberg, H.; Fries, J.; Wu, S.; and Ré, C. 2017. Snorkel: Rapid Training Data Creation with Weak Supervision. In *Proceedings of the VLDB Endowment*, 11(3), 269-282.

Reel, S.; Wong, P.; Wu, B.; Kouadri, S.; and Liu, H. 2018. Identifying tweets from Syria refugees using a Random Forest classifier. In *2018 International Conference on Computational Science and Computational Intelligence*, 1277-1280.

Rettberg, J.W. and Gajjala, R. 2016. Terrorists or cowards: negative portrayals of male Syrian refugees in social media. *Feminist Media Studies*, 16(1), 178-181.

Rijhwani, S.; Zhou, S.; Neubig, G.; and Carbonell, J.G. 2020. Soft Gazetteers for Low-Resource Named Entity Recognition. In *Proceedings of the 58th Annual Meeting of the Association for Computational Linguistics*, 8118-8123.

Riyadi, N.A.E. and Widhiasti, M.R. 2020. Racism and Stereotyping of Refugees: The Use of the Hashtag# Aufschrei by Twitter Users in Germany. In *International University Symposium on Humanities and Arts*, 151-157.

Ross, B.; Rist, M.; Carbonell, G.; Cabrera, B.; Kurowsky, N.; and Wojatzki, M. 2017. Measuring the reliability of hate speech annotations: The case of the European refugee crisis. In *arXiv preprint arXiv:1701.08118*.

Sajir, Z. and Aouragh, M. 2019. Solidarity, social media, and the" refugee crisis": Engagement beyond affect. *International Journal of Communication*, 13, 28.

Sanguinetti, M.; Poletto, F.; Bosco, C.; Patti, V.; and Stranisci, M. 2018. An Italian Twitter corpus of hate speech against immigrants. In Proceedings of the Eleventh International Conference on Language Resources and Evaluation (LREC 2018).

Siapera, E.; Boudourides, M.; Lenis, S.; and Suiter, J. 2018. Refugees and network publics on Twitter: Networked framing, affect, and capture. *Social Media+ Society*, 4(1).

Udwan, G.; Leurs, K.; and Alencar, A. 2020. Digital resilience tactics of Syrian refugees in the Netherlands: Social media for social support, health, and identity. *Social Media+ Society*, 6(2).

Urchs, S.; Wendlinger, L.; Mitrović, J.; and Granitzer, M. 2019. MMoveT15: A Twitter Dataset for Extracting and Analysing Migration-Movement Data of the European Migration Crisis 2015. In *2019 IEEE 28th International Conference on Enabling Technologies: Infrastructure for Collaborative Enterprises*, 146-149.

Vázquez, M.F. and Pérez, F.S. 2019. Hate speech in Spain against Aquarius refugees 2018 in Twitter. In *Proceedings of the Seventh International Conference on Technological Ecosystems for Enhancing Multiculturality*, 906-910.

Volkan, V.D. 2018. Refugees as the other: Large-group identity, terrorism and border psychology. *Group analysis*, 51(3), 343-358.

Waldinger, R., 2018. Immigration and the election of Donald Trump: Why the sociology of migration left us unprepared… and why we should not have been surprised. *Ethnic and Racial Studies*, 41(8), pp.1411-1426.

Waseem, Z. and Hovy, D. 2016. Hateful symbols or hateful people? Predictive features for hate speech detection on Twitter. In *Proceedings of the NAACL student research workshop*, 88-93.

Wolf, T.; Debut, L.; Sanh, V.; Chaumond, J.; Delangue, C.; Moi, A.; Cistac, P.; Rault, T.; Louf, R.; Funtowicz, M.; and Davison, J., 2019. HuggingFace's Transformers: State-of-the-art natural language processing. In *arXiv preprint arXiv:1910.03771*.

Yin, W.; Hay, J.; and Roth, D. 2019. Benchmarking Zero-shot Text Classification: Datasets, Evaluation and Entailment Approach. In Proceedings of the 2019 Conference on EMNLP and the 9th IJCNLP, 3914-3923.

Young, T.; Hazarika, D.; Poria, S.; and Cambria, E. 2018. Recent trends in deep learning based natural language processing. *IEEE Computational Intelligence magazine*, 13(3), 55-75.

Zagheni, E.; Garimella, V.R.K.; Weber, I.; and State, B. 2014. Inferring international and internal migration patterns from Twitter data. In *Proceedings of the 23rd International Conference on World Wide Web (WWW)*, 439-444.

Zagheni, E.; Polimis, K.; Alexander, M.; Weber, I.; and Billari, F.C. 2018. Combining social media data and traditional surveys to nowcast migration stocks. In *Annual Meeting of the Population Association of America*.